\documentclass[10pt,twocolumn,letterpaper]{article}

\usepackage{cvpr}
\usepackage{times}
\usepackage{epsfig}
\usepackage{graphicx}
\usepackage{amsmath}
\usepackage{amssymb}

\graphicspath{{./figures/}}

\usepackage[pagebackref=true,breaklinks=true,letterpaper=true,colorlinks,bookmarks=false]{hyperref}

\cvprfinalcopy 

\def\cvprPaperID{225} 

\ifcvprfinal\fi
\newcommand\blfootnote[1]{%
  \begingroup
  \renewcommand\thefootnote{}\footnote{#1}%
  \addtocounter{footnote}{-1}%
  \endgroup
}

\begin{document}

\title{Object Detection in Videos with Tubelet Proposal Networks}

\author{Kai Kang$^{1,2}$\ \ \ Hongsheng Li$^{2,\star}$\ \ \ Tong Xiao$^{1,2}$\ \ \ Wanli Ouyang$^{2,5}$\ \ \ Junjie Yan$^3$\ \ \ Xihui Liu$^4$\ \ \ Xiaogang Wang$^{2,\star}$\\
\small$^1$Shenzhen Key Lab of Comp. Vis. \& Pat. Rec., Shenzhen Institutes of Advanced Technology, CAS, China\\
\small$^2$The Chinese University of Hong Kong\ \ \ $^3$SenseTime Group Limited\ \ \ \small$^4$Tsinghua University\ \ \ $^5$The University  of Sydney\\
{\tt \{kkang,hsli,xiaotong,wlouyang,xgwang\}@ee.cuhk.edu.hk}\\
{\tt yanjunjie@sensetime.com\ \ \ xh-liu13@mails.tsinghua.edu.cn}
}

\maketitle
\blfootnote{$^\star$Corresponding authors}

\begin{abstract}
Object detection in videos has drawn increasing attention recently with the introduction of the large-scale ImageNet VID dataset.
Different from object detection in static images, temporal information in videos is vital for object detection.
To fully utilize temporal information, state-of-the-art methods \cite{kang2016object,kang2016t} are based on spatiotemporal tubelets, which are essentially sequences of associated bounding boxes across time.
However, the existing methods have major limitations in generating tubelets in terms of quality and efficiency.
Motion-based \cite{kang2016t} methods are able to obtain dense tubelets efficiently, but the lengths are generally only several frames, which is not optimal for incorporating long-term temporal information.
Appearance-based \cite{kang2016object} methods, usually involving generic object tracking, could generate long tubelets, but are usually computationally expensive.
In this work, we propose a framework for object detection in videos, which consists of a novel tubelet proposal network to efficiently generate spatiotemporal proposals, and a Long Short-term Memory (LSTM) network that incorporates temporal information from tubelet proposals for achieving high object detection accuracy in videos.
Experiments on the large-scale ImageNet VID dataset demonstrate the effectiveness of the proposed framework for object detection in videos.
\end{abstract}
\section{Introduction}
\label{sec:intro}
The performance of object detection has been significantly improved recently with the emergence of deep neural networks. Novel neural network structures, such as GoogLeNet \cite{googlenet}, VGG \cite{vgg2014simonyan} and ResNet \cite{he2015deep}, were proposed to improve the learning capability on large-scale computer vision datasets for various computer vision tasks, such as object detection \cite{girshick2015fast,ren2015faster,redmon2015you,ouyang2015deepid}, semantic segmentation \cite{long2015fully,Chen:2015deeplab,kang2014fully}, tracking \cite{wang2015visual,bae2014robust,zhu2014crowd}, scene understanding \cite{shao2015deeply,shao2016slicing,li2017vip}, person search \cite{li2017person,xiao2017joint}, \etc. State-of-the-art object detection frameworks for static images are based on these networks and consist of three main stages \cite{girshick2014rich}. Bounding box proposals are first generated from the input image based on how likely each location contains an object of interest. The appearance features are then extracted from each box proposal to classify them as one of the object classes. Such bounding boxes and their associated class scores are refined by post-processing techniques (\eg, Non-Maximal Suppression) to obtain the final detection results. Multiple frameworks, such as Fast R-CNN \cite{girshick2015fast} and Faster R-CNN \cite{ren2015faster}, followed this research direction and eventually formulated the object detection problem as training end-to-end deep neural networks.

\begin{figure}[tb]
    \centering
    \includegraphics[width=\linewidth]{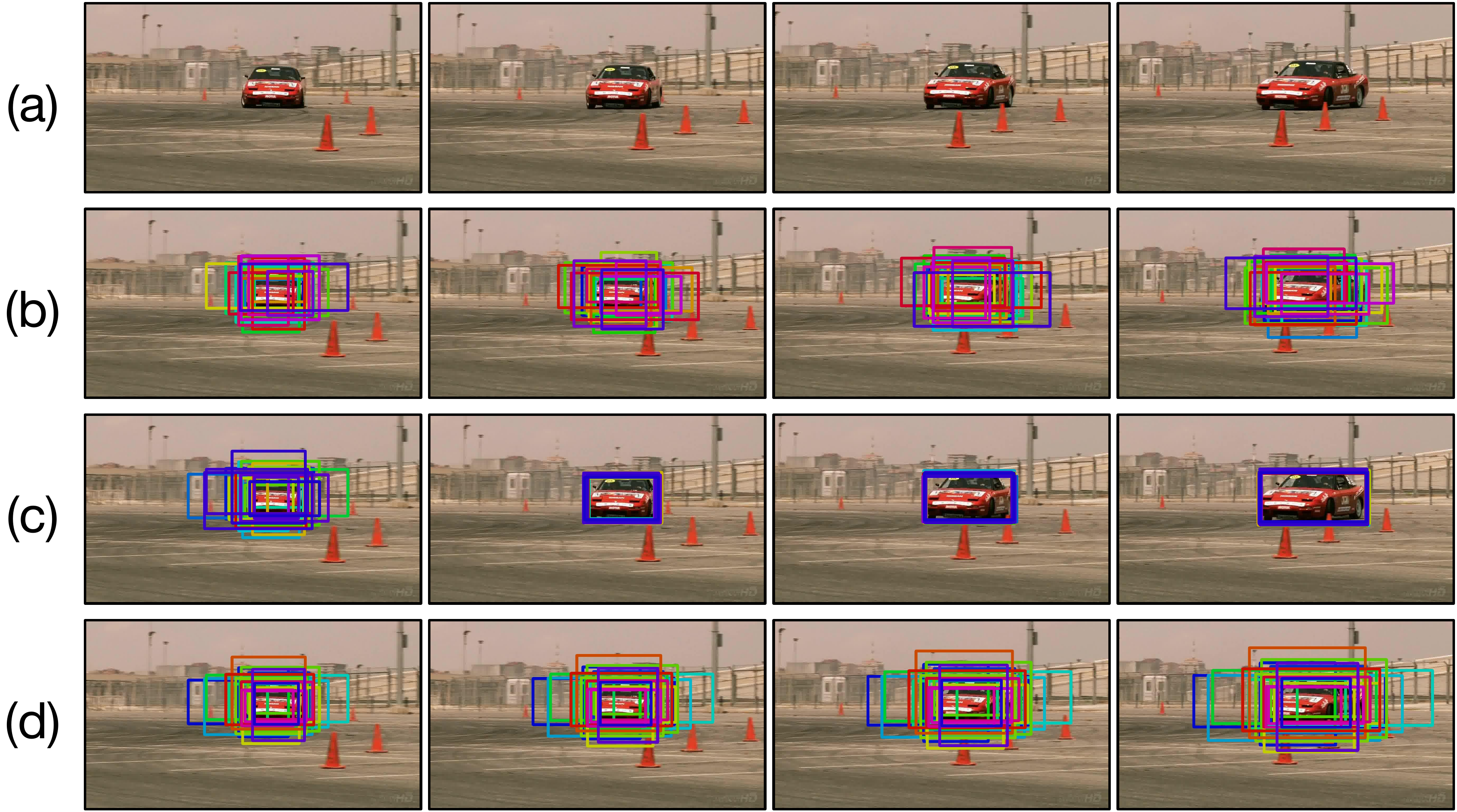}
    \caption{Proposals methods for video object detection. (a) original frames. (b) static proposals have no temporal association, which is hard to incorporate temporal information for proposal classification. (c) bounding box regression methods would focus on the dominant object, lose proposal diversity and may also cause recall drop since all proposals tend to aggregate on the dominant objects. (d) the ideal proposals should have temporal association and have the same motion patterns with the objects while keeping their diversity.}
    \label{fig:motivation}
\end{figure}

Although great success has been achieved in detecting objects on static images, video object detection remains a challenging problem.
Several factors contribute to the difficulty of this problem, which include the drastic appearance and scale changes of the same object over time, object-to-object occlusions, motion blur, and the mismatch between the static-image data and video data.
The new task of detecting objects in videos (VID) introduced by the ImageNet challenge in 2015 provides a large-scale video dataset, which requires labeling every object of 30 classes in each frame of the videos.
Driven by this new dataset, multiple systems \cite{han2016seq,kang2016t,kang2016object} were proposed to extend static-image object detectors for videos.

Similar to the bounding box proposals in the static object detection, the counterpart in videos are called tubelets, which are essentially sequences of bounding boxes proposals.
State-of-the-art algorithms for video object detection utilize the tubelets to some extend to incorporate temporal information for obtaining detection results.
However, the tubelet generation is usually based on the frame-by-frame detection results, which is extremely time consuming.
For instance, the tracking algorithm used by \cite{kang2016t,kang2016object} needs $0.5$ second to process each detection box in each frame, which prevents the systems to generate enough tubelet proposals for classification in an allowable amount of time, since the video usually contains hundreds of frames with hundreds of detection boxes on each frame.
Motion-based methods, such as optical-flow-guided propagation~\cite{kang2016t}, can generate dense tubelets efficiently, but the lengths are usually limited to only several frames (\eg, $7$ frames in \cite{kang2016t}) because of their inconsistent performance for long-term tracking.
The ideal tubelets for video object detection should be long enough to incorporate temporal information while diverse enough to ensure high recall rates (Figure~\ref{fig:motivation}).

To mitigate the problems, we propose a framework for object detection in videos. It consists of a Tubelet Proposal Network (TPN) that simultaneously obtains hundreds of diverse tubelets starting from static proposals, and a Long Short-Term Memory (LSTM) sub-network for estimating object confidences based on temporal information from the tubelets.
Our TPN can efficiently generate tubelet proposals via feature map pooling.
Given a static box proposal at a starting frame, we pool features from the same box locations across multiple frames to train an efficient multi-frame regression neural network as the TPN. It is able to learn complex motion patterns of the foreground objects to generate robust tubelet proposals.
Hundreds of proposals in a video can be tracked simultaneously.
Such tubelet proposals are shown to be of better quality than the ones obtained on each frame independently, which  demonstrates the importance of temporal information in videos.
The visual features extracted from the tubelet boxes are automatically aligned into feature sequences and are suitable for learning temporal features with the following LSTM network, which is able to capture long-term temporal dependency for accurate proposal classification.

The contribution of this paper is that we propose a new deep learning framework that combines tubelet proposal generation and temporal classification with visual-temporal features.
An efficient tubelet proposal generation algorithm is developed to generate tubelet proposals that capture spatiotemporal locations of objects in videos.
A temporal LSTM model is adopted for classifying tubelet proposals with both visual features and temporal features. Such high-level temporal features are generally ignored by existing detection systems but are crucial for object detection in videos.

\section{Related work}
\label{sec:related}
{\bf Object detection in static images.} State-of-the-art object detection systems are all based on deep CNNs. Girshick \etal \cite{girshick2014rich} proposed the R-CNN to decompose the object detection problem into multiple stages including region proposal generation, CNN finetuning, and region classification.
To accelerate the training process of R-CNN, Fast R-CNN \cite{girshick2015fast} was proposed to avoid time-consumingly feeding each image patch from bounding box proposals into CNN to obtain feature representations.  Features of multiple bounding boxes within the same image are warped from the same feature map efficiently via ROI pooling operations. To accelerate the generation of candidate bounding box proposals, Faster R-CNN integrates a Region Proposal Network into the Fast R-CNN framework, and is able to generate box proposals directly with neural networks.

{\bf Object detection in videos.} Since the introduction of the VID task by the ImageNet challenge, there have been multiple object detection systems for detecting objects in videos. These methods focused on post-processing class scores by static-image detectors to enforce temporal consistency of the scores. Han \etal \cite{han2016seq} associated initial detection results into sequences. Weaker class scores along the sequences within the same video were boosted to improve the initial frame-by-frame detection results. Kang \etal \cite{kang2016object} generated new tubelet proposals by applying tracking algorithms to static-image bounding box proposals. The class scores along the tubelet were first evaluated by the static-image object detector and then re-scored by a 1D CNN model. The same group \cite{kang2016t} also tried a different strategy for tubelet classification and re-scoring. In addition, initial detection boxes were propagated to nearby frames according to optical flows between frames, and the class scores not belonging to the top classes were suppressed to enforce temporal consistency of class scores.


{\bf Object localization in videos.} There have been works and datasets \cite{Deselaers:2010localizing,Joulin:2014efficient,Prest:2012learning} on object localization in videos. However, they have a simplified problem setting, where each video is assumed to contain only one known or unknown class and requires annotating only one of the objects in each frame.

\begin{figure}[tb]
    \centering
    \includegraphics[width=\linewidth]{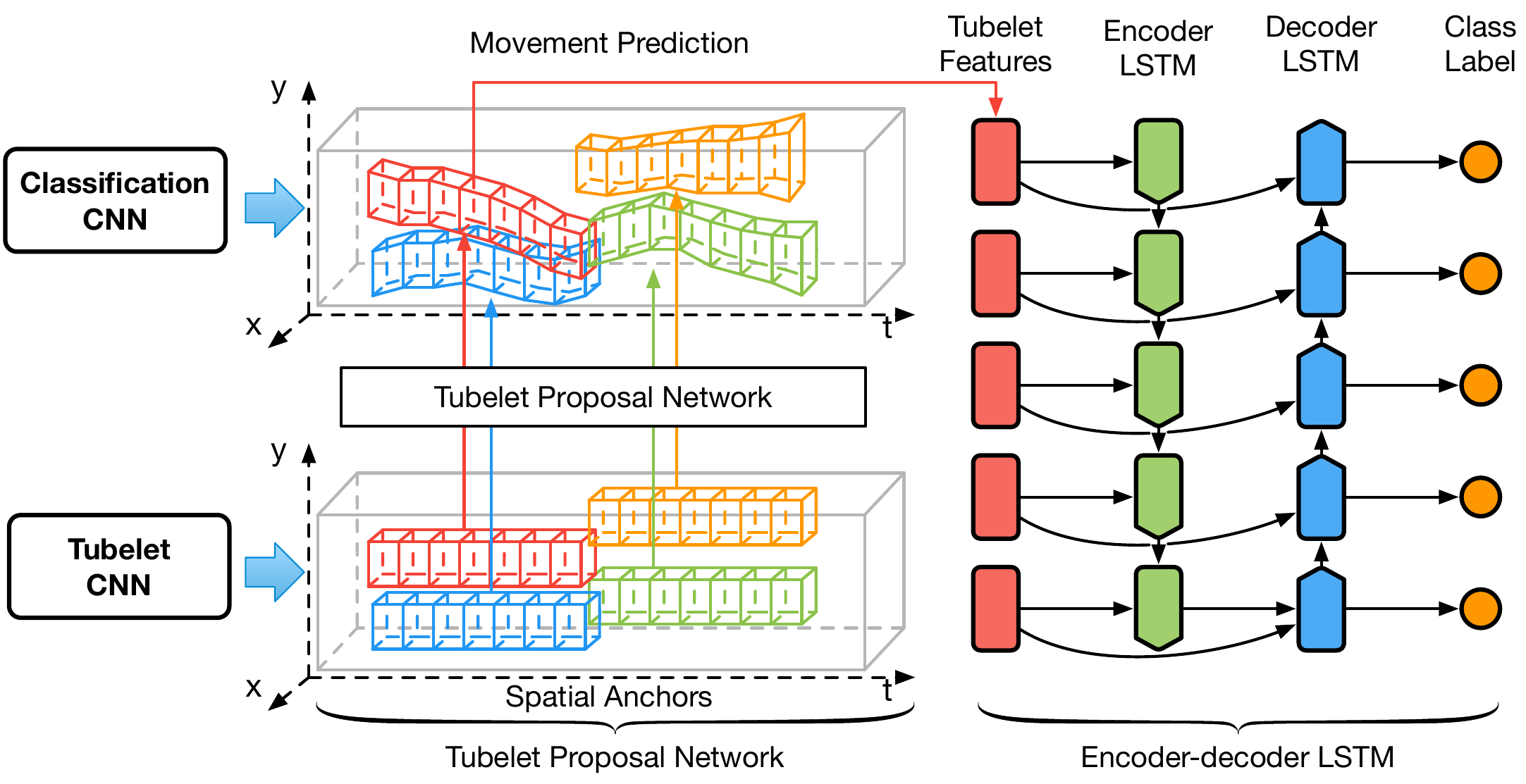}
    \caption{The proposed object detection system, which consists of two main parts. The first is a tubelet proposal network to efficiently generating tubelet proposals. The tubelet proposal network extracts multi-frame features within the spatial anchors, predicts the object motion patterns relative to the spatial anchors and generates tubelet proposals. The gray box indicates the video clip and different colors indicate proposal process of different spatial anchors. The second part is an encoder-decoder CNN-LSTM network to extract tubelet features and classify each proposal boxes into different classes. The tubelet features are first fed into the encoder LSTM by a forward pass to capture the appearance features of the entire sequence. Then the states are copied to the decoder LSTM for a backward pass with the tubelet features. The encoder-decoder LSTM processes the entire clip before outputting class probabilities for each frame.}
    \label{fig:system}
\end{figure}

\section{Tubelet proposal networks} 
\label{sec:tpn}
Existing methods on object detection in videos generates tubelet proposals utilizing either generic single-object tracker starting at a few key frames \cite{kang2016object} or data association methods (\ie tracking-by-detection methods) on per-frame object detection results \cite{han2016seq}.
These methods either are too computationally expensive to generate enough dense tublets, or are likely to drift and result in tracking failures. Even for an 100-fps single-object tracker, it might take about $56$ GPU days to generate tubelets with $300$ bounding boxes per frame for the large-scale ImageNet VID dataset. 

We propose a Tubelet Proposal Network (TPN) which is able to generate tubelet proposals efficiently for videos.
As shown in Figure~\ref{fig:system}, the Tubelet Proposal Network consists of two main components, the first sub-network extracts visual features across time based on static region proposals at a single frame. Our key observation is that, since the receptive fields (RF) of CNNs are generally large enough, we can perform feature map pooling simply at the same bounding box locations across time to extract the visual features of moving objects. Based on the pooled visual features, the second component is a regression layer for estimating bounding boxes' temporal displacements to generate tubelet proposals.

\subsection{Preliminaries on ROI-pooling for regression} 
\label{sub:feature_map_pooling}

There are existing works that utilize feature map pooling for object detection.
The Fast R-CNN framework \cite{girshick2015fast} utilizes ROI-pooling on visual feature maps for object classification and bounding box regression.
The input image is fed into a CNN and forward propagated to generate visual feature maps. Given different object proposals, their visual features are directly ROI-pooled from the feature maps according to the box coordinates. In this way, CNN only needs to forward propagate once for each input image and saves much computational time. Let $b^i_t = (x_t^i, y_t^i, w_t^i, h_t^i)$ denote the $i$th static box proposal at time $t$, where $x$, $y$, $w$ and $h$ represent the two coordinates of the box center, width and height of the box proposal. The ROI-pooling obtains visual features ${\bf r}_t^i \in \mathbb{R}^f$ at box $b^i_t$.

The ROI-pooled features ${\bf r}_t^i$ for each object bounding box proposal can be used for object classification, and, more interestingly, for bounding box regression, which indicates that the visual features obtained by feature map pooling contain necessary information describing objects' locations. Inspired by this technique, we propose to extract multi-frame visual features via ROI-pooling, and use such features for generating tubelet proposals via regression.


\subsection{Static object proposals as spatial anchors}

Static object proposals are class-free bounding boxes indicating the possible locations of objects, which could be efficiently obtained by different proposal methods such as SelectiveSearch \cite{uijlings2013selective}, Edge Boxes \cite{Zitnick:2014edgeboxes} and Region Proposal Networks \cite{ren2015faster}.
For object detection in videos, however, we need both spatial and temporal locations of the objects, which are crucial to incorporate temporal information for accurate object proposal classification.

For general objects in videos, movements are usually complex and difficult to predict. The static object proposals usually have high recall rates (\eg $>$90\%) at individual frames, which is important because it is the upper bound of object detection performance. Therefore, it is natural to use static proposals as starting anchors for estimating their movements at following frames to generate tubelet proposals. If their movements can be robustly estimated, high object recall rate at the following times can be maintained.

Let $b_1^i$ denote a static proposal of interest at time $t=1$. Particularly, to generate a tubelet proposal starting at $b_1^i$, visual features within the $w$-frame temporal window from frame $1$ to $w$ are pooled at the same location $b_1^i$ as ${\bf r}_1^i, {\bf r}_2^i, \cdots, {\bf r}_w^i$ in order to generate the tubelet proposal. We call $b_1^i$ a ``spatial anchor''. The pooled regression features encode visual appearances of the objects. Recovering correspondences between the visual features (${\bf r}_1^i, {\bf r}_2^i, \cdots, {\bf r}_w^i$) leads to accurate tubelet proposals, which is modeled by a regression layer detailed in the next subsection.



The reason why we are able to pool multi-frame features from the same spatial location for tubelet proposals is that CNN feature maps at higher layers usually have large receptive fields. Even if visual features are pooled from a small bounding box, its visual context is far greater than the bounding box itself. Pooling at the same box locations across time is therefore capable of capturing large possible movements of objects.
In Figure~\ref{fig:system}, we illustrate the ``spatial anchors'' for tubelet proposal generation.
The features in the same locations are aligned to predict the movement of the object.

We use a GoogLeNet with Batch Normalization (BN) model \cite{ioffe2015batch} for the TPN.
In our settings, the ROI-pooling layer is connected to ``inception\_4d'' of the BN model, which has a receptive field of $363$ pixels. Therefore, the network is able to handle up to $363$-pixel movement when ROI-pooling the same box locations across time, which is more than enough to capture short-term object movements.
Each static proposal is regarded as an anchor point for feature extraction within a temporal window $w$.



\subsection{Supervisions for tubelet proposal generation} 
\label{sub:movement_loss}

Our goal is to generate tubelet proposals that have high object recall rates at each frame and can accurately track objects. Based on the pooled visual features ${\bf r}_1^i, {\bf r}_2^i, \cdots, {\bf r}_w^i$ at box locations $b_t^i$, we train a regression network $R(\cdot)$ that effectively estimates the relative movements w.r.t. the spatial anchors,
\begin{align}
    m_1^i, m_2^i, \cdots, m_w^i = R({\bf r}_1^i, {\bf r}_2^i, \cdots, {\bf r}_w^i),
\end{align}
where the relative movements $m_t^i$ $=$$(\Delta x^i_t,$ $\Delta y^i_t,$ $\Delta w^i_t,$ $\Delta h^i_t)$ are calculated as
\begin{align}
\Delta x^i_t &= (x^i_t - x^i_1)/w^i_1, \ \ \Delta y^i_t = (y^i_t - y^i_1)/h^i_1, \label{eq:relative_movement} \\
\Delta w^i_t &= \log(w^i_t/w^i_1), \ \ \ \ \ \Delta h^i_t = \log(h^i_t/h^i_1). \nonumber
\end{align}
Once we obtain such relative movements, the actual box locations of the tubelet could be easily inferred.
We adopt a fully-connected layer that takes the concatenated visual features $[{\bf r}_1^i, {\bf r}_2^i, \cdots, {\bf r}_w^i]^T$ as the input, and outputs $4w$ movement values of a tubelet proposal by
\begin{align}
    [m_1^i, \cdots, m_w^i]^T = W_w [{\bf r}_1^i, \cdots, {\bf r}_w^i]^T + b_w,
\end{align}
where $W_w \in \mathbb{R}^{fw \times 4w}$ and $b_w \in \mathbb{R}^{4w}$ are the learnable parameters of the layer.

The remaining problem is how to design proper supervisions for learning the relative movements.
Our key assumption is that the tubelet proposals should have consistent movement patterns with the ground-truth objects. However, given static object proposals as the starting boxes for tubelet generation, they usually do not have a perfect $100\%$ Intersection-over-Union (IoU) ratio with the ground truth object boxes. Therefore, we require static box proposals that are close to ground truth boxes to follow the movement patterns of the ground truth boxes. More specifically, if a static object proposal $b^i_t$ has a greater-than-$0.5$ IoU value with a ground truth box $\hat{b}^i_t$, and the IoU value is greater than those of other ground truth boxes,
our regression layer tries to generate tubelet boxes following the same movement patterns $\hat{m}_t^i$ of the ground truth $\hat{b}^i_t$ as much as possible.
The relative movement targets $\hat{m}_t^i = (\hat x^i_t,\hat y^i_t,\hat w^i_t,\hat h^i_t)$ can be defined w.r.t. the ground truth boxes at time $1$, $\hat{b}_t^1$, in the similar way as Eq. (\ref{eq:relative_movement}).
It is trivial to see that $\hat{m}^i_1=(0,0,0,0).$ Therefore, we only need to predict $\hat{m}_2^i$ to $\hat{m}_w^i$. Note that by learning relative movements w.r.t to the spatial anchors at the first frame, we can avoid cumulative errors in conventional tracking algorithms to some extend.

The movement targets are normalized by their mean $\overline{m_t}$ and standard deviation $\sigma_t$ as the regression objectives, 
\begin{equation}
    \tilde m^i_t = (\hat m^i_t - \overline{m_t}) / \sigma_t, \ \ \ \text{for } t=1,\dots, w.
\end{equation}

To generate $N$ tubelets that follow movement patterns of their associated ground truth boxes, we minimize the following object function w.r.t. all $x^i_t$, $y^i_t$, $w^i_t$, $h^i_t$,
\begin{equation}
    L(\{\tilde{M}\}, \{M\}) = \frac{1}{N}\sum_{i=1}^N \sum_{t=1}^w\sum_{k\in\{x,y,w,h\}} d(\Delta k^i_t),
\end{equation}
where $\{\tilde{M}\}$ and $\{M\}$ are the sets of all normalized movement targets and network outputs, and
\begin{equation}
    d(x)=\begin{cases}
        0.5 x^2 & \text{if } |x| < 1, \\
        |x| - 0.5 & \text{otherwise.}
    \end{cases}
    \label{eq:smooth_l1}
\end{equation}
is the smoothed $L_1$ loss for robust box regression in \cite{girshick2015fast}.

The network outputs $\dot m^i_t$ are mapped back to the real relative movements $m^i_t$ by
\begin{equation}
    m^i_t=(\dot m^i_t + \overline{m_t}) * \sigma_t.
\end{equation}

By our definition, if a static object proposal covers some area the object, it should cover the same portion of object in the later frames (see Figure~\ref{fig:motivation} (d) for examples).

\subsection{Initialization for multi-frame regression layer} 
\label{sub:tpn_init}
The size of the temporal window is also a key factor in the TPN.
The simplest model is a $2$-frame model.
For a given frame, the features within the spatial anchors on current frame and the next frames are extracted and concatenated, $[{\bf r}_1^i, {\bf r}_2^i]^T$, to estimate the movements of $b_1^i$ on the next frames.
However, since the $2$-frame model only utilizes minimal temporal information within a very short temporal window, the generated tubelets may be non-smooth and easy to drift.
Increasing the temporal window utilizes more temporal information so as to estimate more complex movement patterns.

Given the temporal window size $w$, the dimension of the extracted features are $f w$, where $f$ is the dimension of visual features in a single frame within the spatial anchors (\eg, 1024-dimensional ``inception\_5b'' features from the BN model in our settings).
Therefore, the parameter size of the regress layer is of $\mathbb{R}^{fw \times 4w}$ and grows quadratically with the temporal window size $w$.


If the temporal window size is large, randomly initializing such a large matrix has difficulty in learning a good regression layer.
We propose a ``block'' initialization method to use the learned features from $2$-frame model to initialize the multi-frame models.

\begin{figure}[tb]
    \centering
    \includegraphics[width=0.85\linewidth]{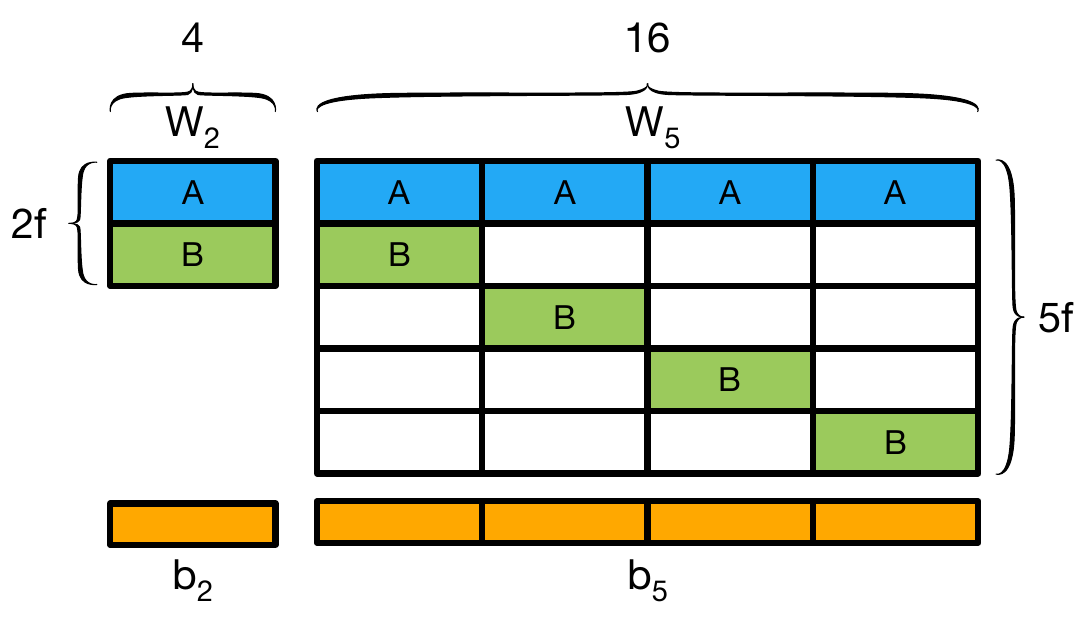}
    \caption{Illustration of the ``block'' initialization method. The $2$-frame model's regression layer has weights $W_2$ and bias $b_2$, the $W_2$ consists of two sub-matrices $A$ and $B$ corresponding to the features of the first and second frames. Then a $5$-frame model's regression layer can be initialized with the sub-matrices as shown in the figure. The bias term $b_5$ is a simple repetition of $b_2$.}
    \label{fig:model_surgery}
\end{figure}
In Figure~\ref{fig:model_surgery}, we show how to use a pre-trained $2$-frame model's regression layer to initialize that of a $5$-frame model.
Since the target $\hat{m}^i_1$ in Equation~(\ref{eq:relative_movement}) is always $(0,0,0,0)$ we only need to estimate movements for the later frames.
The parameter matrix $W_2$ is of size $\mathbb{R}^{2f \times 4}$ since the input features are concatenations of two frames and the bias term $b_2$ is of size $\mathbb{R}^4$.
For the $5$-frame regression layer, the parameter matrix $W_5$ is of size $\mathbb{R}^{5f \times (4\times4)}$ and the bias term $b_5$ is of $\mathbb{R}^{(4\times4)}$.
Essentially, we utilize visual features from frame $1$ \& $2$ to estimate movements in frame $2$, frame $1$ \& $3$ for frame $3$, and so on.
The matrix $W_2$ is therefore divided into two sub-matrices $A \in \mathbb{R}^{f\times 4}$ and $B \in \mathbb{R}^{f\times4}$ to fill the corresponding entries of matrix $W_5$.
The bias term $b_5$ is a repetition of $b_2$ for $4$ times.

In our experiments, we first train a $2$-frame model with random initialization and then use the $2$-frame model to initialize the multi-frame regression layer.

\section{Overall detection framework with tubelet generation and tubelet classification} 
\label{sec:framework}
Based on the Tubelet Proposal Networks, we propose a framework that is efficient for object detection in videos. Compared with state-of-the-art single object tracker, It only takes our TPN $9$ GPU days to generate dense tubelet proposals on the ImageNet VID dataset. It is also capable of utilizing useful temporal information from tubelet proposals to increase detection accuracy.
As shown in Figure~\ref{fig:system}, the framework consists of two networks, the first one is the TPN for generating candidate object tubelets, and the second network is a CNN-LSTM classification network that classifies each bounding box on the tubelets into different object categories.

\subsection{Efficient tubelet proposal generation} 
\label{sub:efficent_tpn}
The TPN is able to estimate movements of each static object proposal within a temporal window $w$.
For object detection in videos in large-scale datasets, we need to not only efficiently generate tubelets for hundreds of spatial anchors in parallel, but also generate tubelets with sufficient lengths to incorporate enough temporal information.

\begin{figure}[tb]
    \centering
    \includegraphics[width=0.85\linewidth]{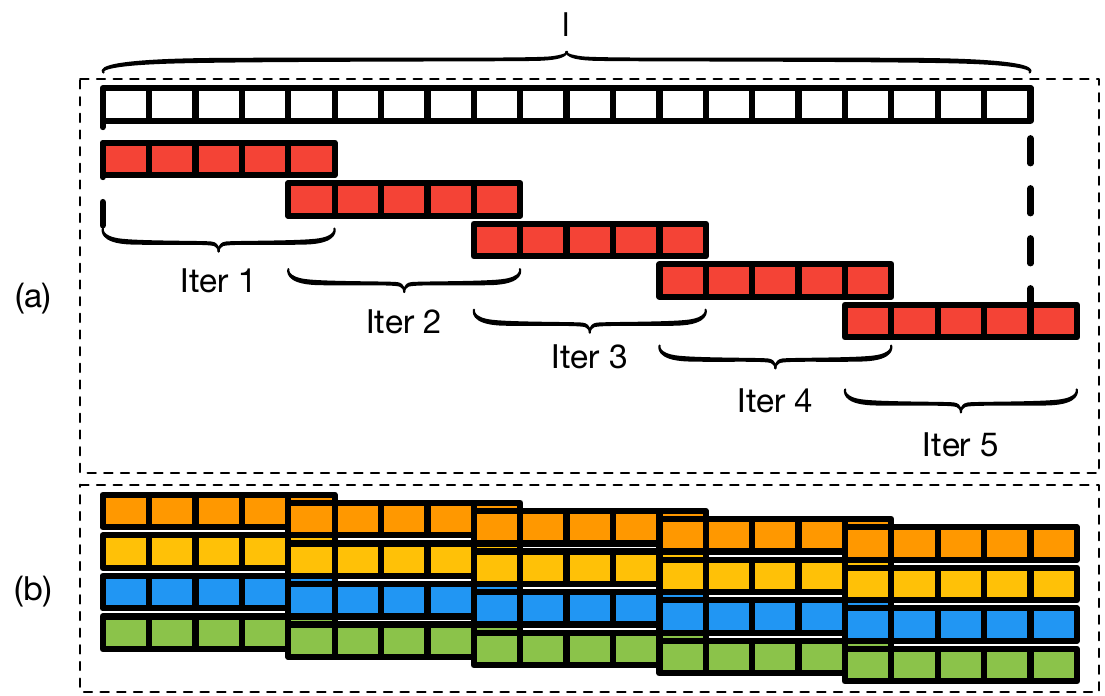}
    \caption{Efficiently generating tubelet proposals. (a) the TPN generates the tubelet proposal of temporal window $w$ and uses the last-frame output of the proposal as static anchors for the next iteration. This process iterates until the whole track length is covered. (b) multiple static anchors in a frame are fed to the Fast R-CNN network with a single forward pass for simultaneously generating multiple tubelet proposals. Different colors indicate different spatial anchors}
    \label{fig:efficient_tpn}
\end{figure}

To generate tubelets with length of $l$, (see illustration in Figure~\ref{fig:efficient_tpn} (a)), we utilize static object proposals on the first frame as spatial anchors, and then iteratively apply TPN with temporal window $w$ until the tubelets cover all $l$ frames.
The last estimated locations of the previous iteration are used as spatial anchors for the next iteration. This process can iterate to generate tubelet proposals of arbitrary lengths.

For $N$ static object proposals in the same starting frame, the bottom CNN only needs to conduct an one-time forward propagation to obtain the visual feature maps, and thus enables efficient generation of hundreds of tubelet proposals (see Figure~\ref{fig:efficient_tpn} (b)).

Compared to previous methods that adopt generic single object trackers, our proposed methods is dramatically faster for generating a large number of tubelets.
The tracking method used in \cite{kang2016object} has reported $0.5$ fps running speed for a single object.
For a typical frame with $300$ spatial anchors, it takes $150$s for each frame.
Our method has an average speed of $0.488$s for each frame, which is about $300\times$ faster.
Even compared to the recent $100$ fps single object tracker in \cite{Held:2016learning}, our method is about $6.14\times$ faster.

\subsection{Encoder-decoder LSTM (ED-LSTM) for temporal classification} 
\label{sub:encoder_decoder}
After generating the length-$l$ tubelet proposal, visual features ${\bf u}_t^1, \cdots, {\bf u}_t^i, \cdots, {\bf u}_{l}^i$ can be pooled from tubelet box locations for object classification with temporal information.
Existing methods \cite{kang2016object,han2016seq,kang2016t} mainly use temporal information in post processing, either propagating detections to neighboring frames or temporally smoothing detection scores.
The temporal consistency of detection results is important, but to capture the complex appearance changes in the tubelets, we need to learn discriminative spatiotemporal features at the tubelet box locations.

As shown in Figure \ref{fig:system}, the proposed classification subnetwork contains a CNN that processes input images to obtain feature maps. Classification features ROI-pooled from each tubelet proposal across time are then fed into a one-layer Long Short-Term Memory (LSTM) network \cite{hochreiter1997long} for tubelet classification. It  is a special type of recurrent neural network (RNN) and is widely investigated for learning spatiotemporal features in recent years.
Each LSTM unit has a memory unit that conveys visual information across the time for incorporating temporal information.

The input for each time step $t$ of the LSTM for the $i$th tubelet are the cell state $c^i_{t-1}$, hidden state $h^i_{t-1}$ of the previous frame, and the classification features ${\bf u}^i_{t}$ pooled at the current time $t$.
The starting state $(c^i_0, h^i_0)$ of the LSTM is set to zeros.
The output is the hidden states $h^i_t$, which is connected to a fully-connected layer for predicting class confidences and another FC layer for box regression.
One problem with the vanilla LSTM is that the initial state may dramatically influence the classification of the first several frames.
Inspired by sequence-to-sequence LSTM in \cite{sutskever2014sequence}, we propose an encoder-decoder LSTM model for object detection in videos as shown in Figure~\ref{fig:system}.
The input features are first fed into an encoder LSTM to encode the appearance features of the entire tubelet into the memory.
The memory and hidden states are then fed into the decoder LSTM, which then classifies the tubelet in the reverse order with the reversed inputs from the last frame back to the first frame.
In this way, better classification accuracy can be achieved by utilizing both past and future information. The low prediction confidences caused by the all-zero initial memory states can be avoided.

\section{Experiments} 
\label{sec:experiments}
\subsection{Datasets and evaluation metrics} 
\label{sub:dataset}
The proposed framework is evaluated on the ImageNet object detection from video (VID) dataset introduced in the ILSVRC 2015 challenge.
There are $30$ object classes in the dataset.
The dataset is split into three subsets: the training set that contains $3862$ videos, the validation set that contains $555$ videos, and the test set that contains $937$ videos.
Objects of the $30$ classes are labeled with ground truth bounding boxes on all the video frames.
Since the ground truth labels for the test set are not publicly available, we report all results on the validation set as a common practice on the ImageNet detection tasks.
The mean average precision (Mean AP) of $30$ classes is used as the evaluation metric.

In addition, we also evaluate our system on the YouTubeObjects (YTO)~\cite{Prest:2012learning} dataset for the object localization task.
The YTO dataset has $10$ object classes, which are a subset of the ImageNet VID dataset.
The YTO dataset is weakly annotated with only one object of one ground truth class in the video.
We only use this dataset for evaluation and the evaluation metric is CorLoc performance measure used in \cite{Deselaers:2010localizing}, \ie, the recall rate of ground-truth boxes with IoU above $0.5$.

\subsection{Base CNN model training} 
\label{sub:static_image}
We choose GoogLeNet with Batch Normalization (BN) \cite{ioffe2015batch} as our base CNN models for both our TPN and CNN-LSTM models without sharing weights between them. The BN model is pre-trained with the ImageNet classification data and fine-tuned on the ImageNet VID dataset.
The static object proposals are generated by a RPN network trained on the ImageNet VID dataset.
The recall rate of the per-frame RPN proposals on the VID validation set is $95.92$ with $300$ boxes on each frame.


To integrate with Fast RCNN framework, we placed the ROI-pooling layer after ``inception\_4d'' rather than the last inception module (``inception\_5b''), because ``inception\_5b'' has $32\times$ down-sampling with a receptive field of $715$ pixels, which is too large for ROI-pooling to generate discriminative features. 
The output size of ROI-pooling is $14\times14$ and we keep the later inception modules and the final global pooling after ``inception\_5b''. We then add one more FC layer for different tasks including tubelet proposal, classification or bounding box regression.

The BN model is trained on $4$ Titan X GPUs for $200,000$ iterations, with $32$ RoIs from $2$ images on each card in every iteration.
The initial learning rate is $5\times 10^{-4}$ and decreases to $1/10$ of its previous value for every $60,000$ iterations.
All BN layers are frozen during the fine-tuning.
After fine-tuning on DET data, the BN model achieves $50.3\%$ mean AP on the ImageNet DET data. After fine-tuning the BN model on the VID data with the same hyper-parameter setting for $90,000$ iterations, it achieves $63.0\%$ mean AP on the VID validation set.

\subsection{TPN training and evaluation} 
\label{sub:tpn_training}
With the fine-tuned BN model, we first train a $2$-frame model on the ImageNet VID dataset.
Since the TPN needs to estimate the movement of the object proposals according ground-truth objects' movements, we only select static proposals that have greater-than-$0.5$ IoU overlaps with ground-truth annotations as spatial anchors following Section \ref{sub:movement_loss}.
For those proposals that do not have greater-than-$0.5$ overlaps with ground-truth boxes, they are not used for training the TPN.
During the test stage, however, all static object proposals in every $20$ frames are used as spatial anchors for tubelet proposal generation. All tubelets are $20$-frame long. The ones starting from negative static proposals are likely to stay in the background regions, or track the foreground objects when they appear in their nearby regions.

We investigate different temporal window sizes $w$ and initialization methods described in Section \ref{sub:tpn_init}. Since the ground truth movements $\hat{m}_t^i$ can be obtained from the ground truth annotations, each positive static proposal has an ``ideal'' tubelet proposal in comply with its associated ground-truth's movements.
Three metrics are used to evaluate the accuracy of generated tubelets by different models (Table \ref{tab:tpn_window}). One is the mean absolute pixel difference (MAD) of the predicted coordinates and their ground truth. The second one is the mean relative pixel difference (MRD) with $x$ differences normalized by widths and $y$ differences normalized by heights.
The third metric is the mean intersection-over-union (IOU) between predicted boxes and target boxes.
From the table, we can see that the $2$-frame baseline model has a MAD of $15.50$, MRD of $0.0730$ and Mean IOU of $0.7966$. For the $5$-frame model, if we initialize the fully-connected regression layer randomly without using the initialization technique (other layers are still initialized by the finetuned BN model), the performance drops significantly compared to that of the $2$-frame model.
The reason might be that the parameter size of the $5$-frame model increases by $10$ times (as shown in Figure~\ref{fig:model_surgery}), which makes it more difficult to train without a good initial point.
However, with the proposed technique, the multi-frame regression layer with the $2$-frame model, the generated tubelets have better accuracy than the $2$-frame model because of the larger temporal context.
\begin{table}[tb]
    \centering
    \footnotesize
    \begin{tabular}{l|c|c|c|c}
    \hline
    \hline
    Method & Window & MAD & MRD & Mean IOU \\
    \hline
    MoveTubelets Random & 2 & 15.50 &  0.0730 & 0.7966\\
    MoveTubelets Random & 5 & 26.00 &  0.1319 & 0.6972\\
    MoveTubelets RNN & 5 & 13.87 & 0.0683 & 0.8060\\
    \hline
    MoveTubelets Block & 5 & \textbf{12.98} &  \textbf{0.0616} & \textbf{0.8244}\\
    MoveTubelets Block & 11 & 15.20 &  0.0761 & 0.8017\\
    MoveTubelets Block & 20 & 18.03 &  0.0874 & 0.7731\\
    \hline
    \hline
    \end{tabular}
    \caption{Evaluation of tubelet proposals obtained with varying window sizes and different initialization methods. As the parameter size grows quadratically with the temporal window. The $5$-frame model with random initialization has much worse accuracy compared to the proposed transformation initialization. As the temporal window grows, the motion pattern becomes more complex and the movement displacement may also exceed the receptive field, which also causes accuracy decreases.}
    \label{tab:tpn_window}
\end{table}

If the temporal window continues to increase, even with the proposed initialization techniques, the performance decreases. This might be because if the temporal window is too large, the movement of the objects might be too complex for the TPN to recover the visual correspondences between far-away frames.
In the later experiments, we use the $5$-frame TPN to generate $20$-frame-long tubelet proposals.

In comparison with our proposed method, an RNN baseline with is implemented by replacing the tubelet regression layer with an RNN layer of $1024$ hidden neurons and a regression layer to predict $4$ motion targets.
As shown in Table~\ref{tab:tpn_window}, the RNN baseline performs worse than our method.

\subsection{LSTM Training} 
\label{sub:lstm_training}
After generating the tubelet proposals, the proposed CNN-LSTM models extract classification features ${\bf u}_t^i$ at tubelet box locations with the finetuned BN model.
The dimension of the features at each time step is $1024$.

The LSTM has $1024$ cell units and $1024$ hidden outputs. For each iteration, $128$ tubelets from $4$ videos are randomly chosen to form a mini-batch. The CNN-LSTM is trained using stochastic gradient descent (SGD) optimization with momentum of $0.9$ for $20000$ iterations.
The parameters are initialized with standard deviation of $0.0002$ and the initial learning rate is $0.1$.
For every $2,000$ iteration, the learning rate decreases by a factor of $0.5$.

\subsection{Results} 
\label{sub:results}
\noindent\textbf{Baseline methods.}
The most basic baseline method is Fast R-CNN static detector \cite{girshick2015fast} (denoted as ``Static''), which needs static proposals on every frame and does not involve any temporal information.
This baseline uses static proposals from the same RPN we use and the Fast R-CNN model is the same as our base BN model.
To validate the effectiveness of the tubelet regression targets, we change them into the precise locations of the ground truth on each frame and also generate tubelet proposals (see Figure~\ref{fig:motivation} (c)).
Then we apply a vanilla LSTM on these tubelet proposals and denote the results as ``LocTubelets+LSTM''.
Our tubelet proposal method is denoted as ``MoveTubelets''.
We also compare with a state-of-the-art single-object tracking method \cite{henriques2015high} denoted as ``KCF''.
As for the CNN-LSTM classification part, the baseline methods are the vanilla LSTM (denoted as ``LSTM''), and our proposed encoder-decoder LSTM is denoted as ``ED-LSTM''.
\begin{table*}[t]
\centering
\scriptsize{
\begin{tabular}{p{2.7cm}|p{0.4cm}p{0.4cm}p{0.4cm}p{0.4cm}p{0.4cm}p{0.4cm}p{0.4cm}p{0.4cm}p{0.4cm}p{0.4cm}p{0.4cm}p{0.4cm}p{0.4cm}p{0.4cm}p{0.4cm}p{0.4cm}}
\hline
Method & \rotatebox{60}{airplane} &  \rotatebox{60}{antelope} &  \rotatebox{60}{bear} &  \rotatebox{60}{bike} &  \rotatebox{60}{bird} &  \rotatebox{60}{bus} &  \rotatebox{60}{car} &  \rotatebox{60}{cattle} &  \rotatebox{60}{dog} &  \rotatebox{60}{d\_cat} &  \rotatebox{60}{elephant} &  \rotatebox{60}{fox} &  \rotatebox{60}{g\_panda} &  \rotatebox{60}{hamster} &  \rotatebox{60}{horse} &  \rotatebox{60}{lion} \\
\hline
Static (Fast RCNN) & 0.821 & 0.784 & 0.665 & 0.656 & 0.661 & 0.772 & 0.523 & 0.491 & 0.571 & 0.720 & 0.681 & 0.768 & 0.718 & 0.897 & 0.651 & 0.201\\
MoveTubelets+Fast RCNN & 0.776 & 0.778 & 0.663 & 0.654 & 0.649 & 0.766 & 0.514 & 0.493 & 0.559 & 0.724 & 0.684 & 0.775 & 0.710 & 0.900 & 0.642 & 0.208\\
\hline
LocTubelets+LSTM & 0.759 & 0.783 & 0.660 & 0.646 & 0.682 & \textbf{0.813} & 0.538 & 0.528 & 0.605 & 0.722 & 0.698 & 0.782 & 0.724 & 0.901 & 0.664 & 0.212\\
MoveTubelets+LSTM & 0.839 & \textbf{0.794} & 0.715 & 0.652 & \textbf{0.683} & 0.794 & 0.533 & \textbf{0.615} & 0.608 & 0.765 & 0.705 & \textbf{0.839} & 0.769 & \textbf{0.916} & 0.661 & 0.158\\
MoveTubelets+ED-LSTM & \textbf{0.846} & 0.781 & \textbf{0.720} & \textbf{0.672} & 0.680 & 0.801 & \textbf{0.547} & 0.612 & \textbf{0.616} & \textbf{0.789} & \textbf{0.716} & 0.832 & \textbf{0.781} & 0.915 & \textbf{0.668} & \textbf{0.216}\\
\hline
\hline
Method & \rotatebox{60}{lizard} &  \rotatebox{60}{monkey} &  \rotatebox{60}{motor} &  \rotatebox{60}{rabbit} &  \rotatebox{60}{r\_panda} &  \rotatebox{60}{sheep} &  \rotatebox{60}{snake} &  \rotatebox{60}{squirrel} &  \rotatebox{60}{tiger} &  \rotatebox{60}{train} &  \rotatebox{60}{turtle} &  \rotatebox{60}{watercraft} &  \rotatebox{60}{whale} &  \rotatebox{60}{zebra} &  \rotatebox{60}{mean AP} &  \\
\hline
Static (Fast RCNN) & 0.638 & 0.347 & 0.741 & 0.457 & 0.558 & 0.541 & 0.572 & 0.298 & 0.815 & 0.720 & 0.744 & 0.557 & 0.432 & \textbf{0.894} & 0.630 & \\
MoveTubelets+Fast RCNN & 0.646 & 0.320 & 0.691 & 0.454 & 0.582 & 0.540 & 0.567 & 0.286 & 0.806 & 0.730 & 0.737 & 0.543 & 0.414 & 0.885 & 0.623\\
\hline
LocTubelets+LSTM & 0.743 & 0.334 & 0.727 & 0.513 & 0.555 & 0.613 & \textbf{0.688} & 0.422 & 0.813 & 0.781 & 0.760 & 0.609 & 0.429 & 0.874 & 0.653\\
MoveTubelets+LSTM & \textbf{0.746} & 0.347 & \textbf{0.771} & \textbf{0.525} & \textbf{0.710} & 0.609 & 0.637 & 0.406 & 0.845 & \textbf{0.786} & \textbf{0.774} & 0.602 & 0.637 & 0.890 & 0.678\\
MoveTubelets+ED-LSTM & 0.744 & \textbf{0.366} & 0.763 & 0.514 & 0.706 & \textbf{0.642} & 0.612 & \textbf{0.423} & \textbf{0.848} & 0.781 & 0.772 & \textbf{0.615} & \textbf{0.669} & 0.885 & \textbf{0.684}\\
\hline
\end{tabular}
}
\caption{AP list on ImageNet VID validation set by the proposed method and compared methods.}
\label{tab:ap_list}
\end{table*}

\begin{table}[h]
    \centering
    \footnotesize
    \begin{tabular}{l|c}
    \hline
    \hline
    Static (Fast RCNN) & 0.630\\
    \hline
    TCNN \cite{kang2016t} & 0.615\\
    Seq-NMS \cite{han2016seq} & 0.522\\
    Closed-loop \cite{galteri2017spatio} & 0.500\\
    \hline
    KCF Tracker \cite{henriques2015high} + Fast R-CNN & 0.567 \\
    MoveTubelets + Fast R-CNN & 0.623 \\
    \hline
    MoveTubelets+LSTM &  0.678 \\
    MoveTubelets+ED-LSTM (proposed) & \textbf{0.684} \\
    \hline
    \hline
    \end{tabular}
    \caption{Mean AP for baseline models and proposed methods.}
    \label{tab:mean_ap}
\end{table}

\noindent\textbf{Results on ImageNet VID dataset.}
The quantitative results on the ImageNet VID dataset are shown in Table~\ref{tab:ap_list} and \ref{tab:mean_ap}. As a convention of detection tasks on the ImageNet dataset, we report the results on the validation set.
The performance of the baseline Fast R-CNN detector finetuned on the ImageNet VID dataset has a Mean AP of $0.630$ (denoted as ``Static''). Compare to the best single model performance in \cite{kang2016t}, which has a Mean AP of $0.615$ using only the VID data, the baseline detector has an $1.5\%$ performance gain.

Directly applying the baseline static detector on the TPN tubelets with temporal window of $5$ results in a Mean AP of $0.623$ (denoted as ``MoveTubelets+Fast RCNN'').
In comparison, a state-of-the-art tracker \cite{henriques2015high} with the baseline static detector (``KCF+Fast RCNN'') has a Mean AP of only $0.567$.
In addition, although the KCF tracker runs at $50$ fps for single object tracking, it takes $6$ seconds to process one frame with $300$ proposals.
Our method is $12\times$ faster.

Applying the vanilla LSTM on the tubelet proposals increases the Mean AP to $0.678$ (denoted as ``MoveTubelets+LSTM''), which has $5.5\%$ performance gain over the tubelet results and $4.8\%$ increase over the static baseline results.
This shows that the LSTM is able to learn appearance and temporal features from the tubelet proposals to improve the classification accuracy.
Especially for class of ``whale'', the AP has over $25\%$ improvement since whales constantly emerge from the water and submerge. A detector has to observe the whole process to classify them correctly.

Compared to bounding box regression tubelet proposal baseline, our tubelet proposal model has $2.5\%$ improvement which shows that our tubelet proposals have more diversity to incorporate temporal information.
Changing to the encoder-decoder LSTM model has a Mean AP of $0.684$ (denoted as ``MoveTubelets+ED-LSTM'') with a $0.6\%$ performance gain over the vanilla LSTM model with performance increases on over half of the classes.
One thing to notice is that our encoder-decoder LSTM model performs better than or equal to the tubelet baseline results on all the classes, which means that learning the temporal features consistently improves the detection results.

\begin{figure}[tb]
  \centering
  \includegraphics[width=0.95\linewidth]{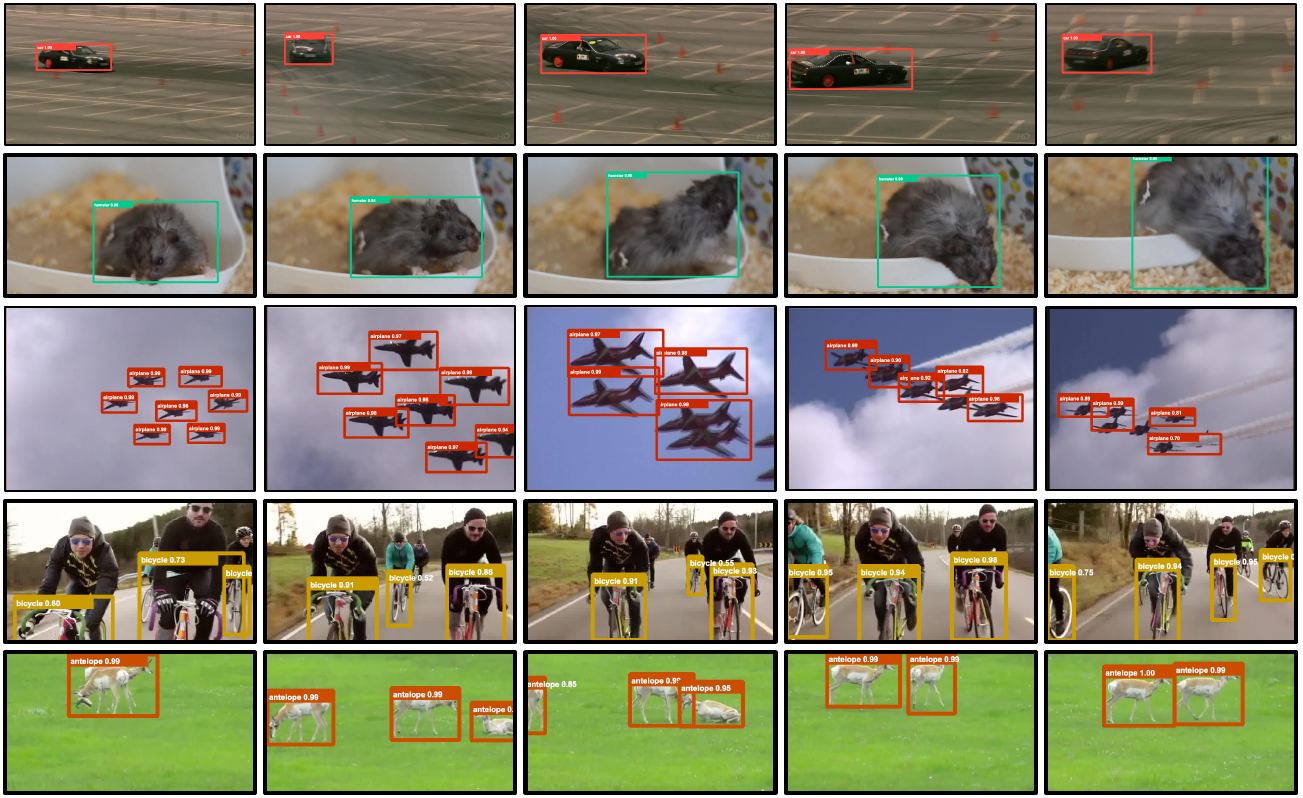}
  \caption{Qualitative results on the ImageNet VID dataset. The bounding boxes are tight and stably concentrate on the objects since the RoIs for each frame are based on the predicted locations on the previous frame. The last $3$ rows show the robustness to handle scenes with multiple objects.)}
  \label{fig:qualitative_results}
\end{figure}

The qualitative results on the ImageNet VID dataset are shown in Figure~\ref{fig:qualitative_results}. The bounding boxes are tight to the objects and we able to track and detect multiple objects during long periods of time.

\noindent\textbf{Localization on the YouTubeObjects dataset.}
In addition to the object detection in video task on the ImageNet VID dataset. We also evaluate our system on video object localization task with the YouTubeObjects (YTO) dataset.

For each test video, we generate tubelet proposals and apply the encoder-decoder LSTM model to classify the tubelet proposals. For each test class, we select the tubelet box with the maximum detection score on the test frames, if the box has over $0.5$ IOU overlap with one of the ground truth boxes, this frame is accurately localized.
The system is trained on the ImageNet VID dataset and is directly applied for testing without any finetuning on the YTO dataset.
\begin{table}[h]
    \centering
    \tiny
    \begin{tabular}{l|p{0.15cm}p{0.15cm}p{0.15cm}p{0.15cm}p{0.15cm}p{0.15cm}p{0.15cm}p{0.15cm}p{0.15cm}p{0.15cm}|c}
    \hline
    \hline
    Method & aero & bird & boat & car & cat & cow & dog & horse & mbike & train & Avg. \\
    \hline
    Prest \etal \cite{Prest:2012learning} & 51.7 & 17.5 & 34.4 & 34.7 & 22.3 & 17.9 & 13.5 & 26.7 & 41.2 & 25.0 & 28.5 \\
    Joulin \etal \cite{Joulin:2014efficient} & 25.1 & 31.2 & 27.8 & 38.5 & 41.2 & 28.4 & 33.9 & 35.6 & 23.1 & 25.0 & 31.0 \\
    Kwak \etal \cite{Kwak:2015unsupervised} & 56.5 & 66.4 & 58.0 & 76.8 & 39.9 & 69.3 & 50.4 & 56.3 & 53.0 & 31.0 & 55.7 \\
    Kang \etal \cite{kang2016object} & \textbf{94.1} & 69.7 & 88.2 & 79.3 & 76.6 & 18.6 & 89.6 & \textbf{89.0} & 87.3 & 75.3 & 76.8 \\
    \hline
    MoveTubelets+ED-LSTM & 91.2 & \textbf{99.4} & \textbf{93.1} & \textbf{94.8} & \textbf{94.3} & \textbf{99.3} & \textbf{90.2} & 87.8 & \textbf{89.7} & \textbf{84.2} & \textbf{92.4} \\
    \hline
    \hline
    \end{tabular}
    \caption{Localization results on the YouTubeObjects dataset. Our model outperforms previous method with large margin.}
    \label{tab:yto_scores}
\end{table}
We compare with several state-of-the-art results on the YTO dataset, and our system outperforms them with a large margin. Compared to the second best results in \cite{kang2016object}, our system has $15.6\%$ improvement.

\section{Conclusion} 
\label{sec:conclusion}
In this work, we propose a system for object detection in videos.
The system consists of a novel tubelet proposal network that efficiently generates tubelet proposals and an encoder-decoder CNN-LSTM model to learn temporal features from the tubelets.
Our system is evaluated on the ImageNet VID dataset for object detection in videos and the YTO dataset for object localization.
Experiments demonstrate the effectiveness of our proposed framework.

\noindent\textbf{Acknowledgments.}
This work is supported in part by SenseTime Group Limited, in part by the General Research Fund through the Research Grants Council of Hong Kong under Grants CUHK14207814, CUHK14206114, CUHK14205615, CUHK14213616, CUHK14203015, CUHK14239816, and CUHK419412, in part by the Hong Kong Innovation and Technology Support Programme Grant ITS/121/15FX, in part by National Natural Science Foundation of China  under Grant 61371192, in part by the Ph.D. Program Foundation of China under Grant 20130185120039, and in part by the China Postdoctoral Science Foundation under Grant 2014M552339.

{\small
\bibliographystyle{ieee}
\bibliography{egbib}

\begin{thebibliography}{10}\itemsep=-1pt

\bibitem{bae2014robust}
S.-H. Bae and K.-J. Yoon.
\newblock Robust online multi-object tracking based on tracklet confidence and
  online discriminative appearance learning.
\newblock {\em CVPR}, 2014.

\bibitem{Chen:2015deeplab}
L.-C. Chen, G.~Papandreou, I.~Kokkinos, K.~Murphy, and A.~L. Yuille.
\newblock {DeepLab: Semantic Image Segmentation with Deep Convolutional Nets,
  Atrous Convolution, and Fully Connected CRFs}.
\newblock In {\em ICLR}, 2015.

\bibitem{Deselaers:2010localizing}
T.~Deselaers, B.~Alexe, and V.~Ferrari.
\newblock {Localizing Objects While Learning Their Appearance}.
\newblock {\em ECCV}, 2010.

\bibitem{galteri2017spatio}
L.~Galteri, L.~Seidenari, M.~Bertini, and A.~Del~Bimbo.
\newblock Spatio-temporal closed-loop object detection.
\newblock {\em TIP}, 2017.

\bibitem{girshick2015fast}
R.~Girshick.
\newblock Fast r-cnn.
\newblock {\em ICCV}, 2015.

\bibitem{girshick2014rich}
R.~Girshick, J.~Donahue, T.~Darrell, and J.~Malik.
\newblock Rich feature hierarchies for accurate object detection and semantic
  segmentation.
\newblock {\em CVPR}, 2014.

\bibitem{han2016seq}
W.~Han, P.~Khorrami, T.~L. Paine, P.~Ramachandran, M.~Babaeizadeh, H.~Shi,
  J.~Li, S.~Yan, and T.~S. Huang.
\newblock Seq-nms for video object detection.
\newblock {\em arXiv preprint arXiv:1602.08465}, 2016.

\bibitem{he2015deep}
K.~He, X.~Zhang, S.~Ren, and J.~Sun.
\newblock Deep residual learning for image recognition.
\newblock {\em arXiv preprint arXiv:1512.03385}, 2015.

\bibitem{Held:2016learning}
D.~Held, S.~Thrun, and S.~Savarese.
\newblock {Learning to Track at 100 FPS with Deep Regression Networks}.
\newblock In {\em ECCV}, 2016.

\bibitem{henriques2015high}
J.~F. Henriques, R.~Caseiro, P.~Martins, and J.~Batista.
\newblock High-speed tracking with kernelized correlation filters.
\newblock {\em TPAMI}, 2015.

\bibitem{hochreiter1997long}
S.~Hochreiter and J.~Schmidhuber.
\newblock Long short-term memory.
\newblock {\em Neural computation}, 9(8):1735--1780, 1997.

\bibitem{ioffe2015batch}
S.~Ioffe and C.~Szegedy.
\newblock Batch normalization: Accelerating deep network training by reducing
  internal covariate shift.
\newblock {\em arXiv preprint arXiv:1502.03167}, 2015.

\bibitem{Joulin:2014efficient}
A.~Joulin, K.~Tang, and L.~Fei-Fei.
\newblock {Efficient Image and Video Co-localization with Frank-Wolfe
  Algorithm}.
\newblock {\em ECCV}, 2014.

\bibitem{kang2016t}
K.~Kang, H.~Li, J.~Yan, X.~Zeng, B.~Yang, T.~Xiao, C.~Zhang, Z.~Wang, R.~Wang,
  X.~Wang, et~al.
\newblock T-cnn: Tubelets with convolutional neural networks for object
  detection from videos.
\newblock {\em arXiv preprint arXiv:1604.02532}, 2016.

\bibitem{kang2016object}
K.~Kang, W.~Ouyang, H.~Li, and X.~Wang.
\newblock Object detection from video tubelets with convolutional neural
  networks.
\newblock In {\em CVPR}, 2016.

\bibitem{kang2014fully}
K.~Kang and X.~Wang.
\newblock Fully convolutional neural networks for crowd segmentation.
\newblock {\em arXiv preprint arXiv:1411.4464}, 2014.

\bibitem{Kwak:2015unsupervised}
S.~Kwak, M.~Cho, I.~Laptev, J.~Ponce, and C.~Schmid.
\newblock {Unsupervised Object Discovery and Tracking in Video Collections}.
\newblock {\em ICCV}, 2015.

\bibitem{li2017person}
S.~Li, T.~Xiao, H.~Li, B.~Zhou, D.~Yue, and X.~Wang.
\newblock Person search with natural language description.
\newblock In {\em CVPR}, 2017.

\bibitem{li2017vip}
Y.~Li, W.~Ouyang, and X.~Wang.
\newblock Vip-cnn: A visual phrase reasoning convolutional neural network for
  visual relationship detection.
\newblock In {\em CVPR}, 2017.

\bibitem{long2015fully}
J.~Long, E.~Shelhamer, and T.~Darrell.
\newblock Fully convolutional networks for semantic segmentation.
\newblock In {\em CVPR}, 2015.

\bibitem{ouyang2015deepid}
W.~Ouyang, X.~Wang, X.~Zeng, S.~Qiu, P.~Luo, Y.~Tian, H.~Li, S.~Yang, Z.~Wang,
  C.-C. Loy, et~al.
\newblock Deep{ID}-net: Deformable deep convolutional neural networks for
  object detection.
\newblock {\em CVPR}, 2015.

\bibitem{Prest:2012learning}
A.~Prest, C.~Leistner, J.~Civera, C.~Schmid, and V.~Ferrari.
\newblock {Learning object class detectors from weakly annotated video}.
\newblock {\em CVPR}, 2012.

\bibitem{redmon2015you}
J.~Redmon, S.~Divvala, R.~Girshick, and A.~Farhadi.
\newblock You only look once: Unified, real-time object detection.
\newblock {\em arXiv preprint arXiv:1506.02640}, 2015.

\bibitem{ren2015faster}
S.~Ren, K.~He, R.~Girshick, and J.~Sun.
\newblock Faster r-cnn: Towards real-time object detection with region proposal
  networks.
\newblock {\em NIPS}, 2015.

\bibitem{shao2015deeply}
J.~Shao, K.~Kang, C.~Change~Loy, and X.~Wang.
\newblock Deeply learned attributes for crowded scene understanding.
\newblock In {\em CVPR}, 2015.

\bibitem{shao2016slicing}
J.~Shao, C.-C. Loy, K.~Kang, and X.~Wang.
\newblock Slicing convolutional neural network for crowd video understanding.
\newblock In {\em CVPR}, 2016.

\bibitem{vgg2014simonyan}
K.~Simonyan and A.~Zisserman.
\newblock Very deep convolutional networks for large-scale image recognition.
\newblock {\em ICLR}, 2015.

\bibitem{sutskever2014sequence}
I.~Sutskever, O.~Vinyals, and Q.~V. Le.
\newblock Sequence to sequence learning with neural networks.
\newblock In {\em NIPS}, 2014.

\bibitem{googlenet}
C.~Szegedy, W.~Liu, Y.~Jia, P.~Sermanet, S.~Reed, D.~Anguelov, D.~Erhan,
  V.~Vanhoucke, and A.~Rabinovich.
\newblock Going deeper with convolutions.
\newblock {\em CVPR}, 2015.

\bibitem{uijlings2013selective}
J.~R. Uijlings, K.~E. van~de Sande, T.~Gevers, and A.~W. Smeulders.
\newblock Selective search for object recognition.
\newblock {\em IJCV}, 2013.

\bibitem{wang2015visual}
L.~Wang, W.~Ouyang, X.~Wang, and H.~Lu.
\newblock Visual tracking with fully convolutional networks.
\newblock {\em ICCV}, 2015.

\bibitem{xiao2017joint}
T.~Xiao, S.~Li, B.~Wang, L.~Lin, and X.~Wang.
\newblock Joint detection and identification feature learning for person
  search.
\newblock In {\em CVPR}, 2017.

\bibitem{zhu2014crowd}
F.~Zhu, X.~Wang, and N.~Yu.
\newblock Crowd tracking with dynamic evolution of group structures.
\newblock In {\em ECCV}, 2014.

\bibitem{Zitnick:2014edgeboxes}
C.~L. Zitnick and P.~Dollar.
\newblock {Edge Boxes: Locating Object Proposals from Edges}.
\newblock {\em ECCV}, 2014.

\end{thebibliography}
}


\end{document}